\documentclass{article}
\usepackage{iclr2020_conference,times}
\usepackage{amsmath,amsfonts,bm}

\def\eqref#1{equation~\ref{#1}}
\def\1{\bm{1}}

\DeclareMathAlphabet{\mathsfit}{\encodingdefault}{\sfdefault}{m}{sl}
\SetMathAlphabet{\mathsfit}{bold}{\encodingdefault}{\sfdefault}{bx}{n}

\usepackage{hyperref}
\usepackage{url}
\usepackage{epsfig}
\usepackage{graphicx}
\usepackage{subcaption}
\usepackage{amsmath}
\usepackage{rotating}
\usepackage{multirow}
\usepackage{caption}

\title{Memory-Based Graph Networks}

\author{Amir Hosein Khasahmadi$^{1,2}$ \thanks{ Work done during internship at Autodesk Toronto AI Lab.}
, Kaveh Hassani$^{3}$, Parsa Moradi$^{4}$, Leo Lee$^{1,2}$, Quaid Morris$^{1,2}$ \vspace{3mm} \\
$^{1}$University of Toronto, Toronto, Canada \qquad\qquad\ \   $^{2}$Vector Institute, Toronto, Canada \\
$^{3}$Autodesk AI Lab, Toronto, Canada \hfill $^{4}$Sharif University of Technology, Tehran, Iran \\
\texttt{\small \{amirhosein,ljlee\}@psi.toronto.edu} \quad\quad \ \ \ 
\texttt{\small kaveh.hassani@autodesk.com} \\
\texttt{\small parsa.mordadi@ee.sharif.edu} \qquad\qquad\quad \quad \ 
\texttt{\small quaid.morris@utoronto.ca}
}

\iclrfinalcopy 
\begin{document}

\maketitle

\begin{abstract}
Graph neural networks (GNNs) are a class of deep models that operate on data with arbitrary topology represented as graphs. We introduce an efficient memory 
layer for GNNs that can jointly learn node representations and coarsen the graph. We also introduce two new networks based on this layer: memory-based GNN 
(MemGNN) and graph memory network (GMN) that can learn hierarchical graph representations. The experimental results show that the proposed models achieve 
state-of-the-art results in eight out of nine graph classification and regression benchmarks. We also show that the learned representations could correspond to 
chemical features in the molecule data. Code and reference implementations are released at: \href{https://github.com/amirkhas/GraphMemoryNet}
{\color{blue}https://github.com/amirkhas/GraphMemoryNet}
\end{abstract}
\section{Introduction}
%------------------------------------------------------------------------------------------------------------------------------------------------------------------------------------------------------------------------
Graph neural networks (GNNs) \citep{wu_2019_arxiv, zhou_2018_arxiv, zhang_2018_arxiv} are a class of deep models that operate on data with arbitrary topology 
represented as graphs such as social networks \citep{kipf_2016_iclr}, knowledge graphs \citep{vivona_2019_arxiv}, molecules \citep{duvenaud_2015_nips}, point clouds 
\citep{hassani_2019_iccv}, and robots \citep{wang_2018_iclr}. Unlike regular-structured inputs such as grids (e.g., images and volumetric data) and sequences (e.g., speech 
and text), GNN inputs are permutation-invariant variable-size graphs consisting of nodes and their interactions. GNNs such as gated GNN (GGNN) 
\citep{li_2015_iclr}, message passing neural network (MPNN) \citep{gilmer_2017_icml}, graph convolutional network (GCN) \citep{kipf_2016_iclr}, and graph attention 
network (GAT) \citep{velickovic_2018_iclr} learn node representations through an iterative process of transferring, transforming, and aggregating the node 
representations from topological neighbors. Each iteration expands the receptive field by one hop and after $k$ iterations the nodes within $k$ hops influence the node 
representations of one another. GNNs are shown to learn better representations compared to random walks \citep{grover_2016_kdd,perozzi_2014_kdd}, matrix 
factorization \citep{belkin_2002_nips, ou_kdd_2016}, kernel methods \citep{shervashidze_2011_jmlr, kriege_2016_nips}, and probabilistic graphical models 
\citep{dai_2016_icml}. 

%------------------------------------------------------------------------------------------------------------------------------------------------------------------------------------------------------------------------
These models, however, cannot learn hierarchical representations as they do not exploit the compositional nature of graphs. Recent work such as differentiable 
pooling (DiffPool) \citep{ying_2018_nips}, TopKPool \citep{gao_2019_icml}, and self-attention graph pooling (SAGPool) \citep{lee_2019_icml} introduce parametric graph 
pooling layers that allow GNNs to learn hierarchical graph representations by stacking interleaved layers of GNN and pooling layers. These layers cluster nodes in 
the latent space. The clusters may correspond to communities in a social network or potent functional groups within a chemical dataset. Nevertheless, these models 
are not efficient as they require an iterative process of message passing after each pooling layer.

%------------------------------------------------------------------------------------------------------------------------------------------------------------------------------------------------------------------------
In this paper, we introduce a memory layer for joint graph representation learning and graph coarsening that consists of a multi-head array of memory keys and a 
convolution operator to aggregate the soft cluster assignments from different heads. The queries to a memory layer are node representations from the previous layer 
and the outputs are the node representations of the coarsened graph. The memory layer does not explicitly require connectivity information and unlike GNNs relies 
on the global information rather than local topology. Hence, it does not struggle with over-smoothing problem \citep{Xu_2018_icml, li_2018_aaai}. These properties 
make memory layers more efficient and improves their performance. We also introduce two networks based on the proposed layer: memory-based GNN 
(MemGNN) and graph memory network (GMN). MemGNN consists of a GNN that learns the initial node representations, and a stack of memory layers that learn 
hierarchical representations up to the global graph representation. GMN, on the other hand, learns the hierarchical representations purely based on memory layers 
and hence does not require message passing. 
%------------------------------------------------------------------------------------------------------------------------------------------------------------------------------------------------------------------------
\section{Related Work}
%------------------------------------------------------------------------------------------------------------------------------------------------------------------------------------------------------------------------
\textbf{Memory augmented neural networks (MANNs)} utilize external memory with differentiable read-write operators allowing them to explicitly access the past 
experiences and are shown to enhance reinforcement learning \citep{pritzel_2017_icml}, meta learning \citep{santoro_2016_icml}, few-shot learning 
\citep{vinyals_2016_nips}, and multi-hop reasoning \citep{weston_2015_iclr}. Unlike RNNs, in which the memory is represented within their hidden states, the decoupled 
memory in MANNs allows them to store and retrieve longer term memories with less parameters. The memory can be implemented as: \textit{key-value memory} such as 
neural episodic control \citep{pritzel_2017_icml} and product-key memory layers \citep{lample_2019_arxiv}, or \textit{array-structured memory} such as neural Turing machine 
(NTM) \citep{graves_2014_arxiv}, prototypical networks \citep{snell_2017_nips}, memory networks \citep{weston_2015_iclr}, and sparse access memory (SAM) 
\citep{rae_2016_nips}. Our memory layer consists of a multi-head array of memory keys.

%------------------------------------------------------------------------------------------------------------------------------------------------------------------------------------------------------------------------
\textbf{Graph neural networks (GNNs)} mostly use message passing to learn node representations over graphs. GraphSAGE \citep{hamilton_2017_nips} learns 
representations by sampling and aggregating neighbor nodes whereas GAT \citep{velickovic_2018_iclr} uses attention mechanism to aggregate representations from all 
neighbors. GCN extend the convolution operator to arbitrary topology. Spectral GCNs \citep{bruna_2014_iclr, defferrard_2016_nips, kipf_2016_iclr} use spectral filters 
over graph Laplacian to define the convolution in the Fourier domain. These models are less efficient compared to spatial GCNs \citep{schlichtkrull_2018_eswc, 
ma_2019_icml} which directly define the convolution on graph patches centered on nodes. Our memory layer uses a feed-forward network to learn the node 
representations.

%------------------------------------------------------------------------------------------------------------------------------------------------------------------------------------------------------------------------
\textbf{Graph pooling} can be defined in global or hierarchical manners. In former, node representations are aggregated into a graph representation by a readout layer 
implemented using \textit{arithmetic operators} such as summation or averaging \citep{hamilton_2017_nips, kipf_2016_iclr} or \textit{set neural networks} such as Set2Set 
\citep{vinyals_2015_iclr} and SortPool \citep{morris_2019_aaai}. In latter, graphs are coarsened in each layer to capture the hierarchical structure. Efficient non-parametric 
methods such as clique pooling \citep{Luzhnica_2019_iclr}, kNN pooling \citep{wang_2018_eccv}, and Graclus \citep{dhillon_2007_tpami} rely on topological information 
but are outperformed by parametric models such as edge contraction pooling \citep{diehl_2019_arxiv}.

%------------------------------------------------------------------------------------------------------------------------------------------------------------------------------------------------------------------------
DiffPool \citep{ying_2018_nips} trains two parallel GNNs to compute node representations and cluster assignments using a multi-term loss including classification, 
link prediction, and entropy losses, whereas Mincut pool \citep{Bianchi_2019_arxiv} trains a sequence of a GNN and an MLP using classification loss and the minimum 
cut objective. TopKPool \citep{cangea_2018_nips, gao_2019_icml} computes node scores by learning projection vectors and dropping all the nodes except the top 
scoring nodes. SAGPool \citep{lee_2019_icml} extends the TopKPool by using graph convolutions to take neighbor node features into account. We use a 
clustering-friendly distribution to compute the attention scores between nodes and clusters.
%------------------------------------------------------------------------------------------------------------------------------------------------------------------------------------------------------------------------
\section{Method}
\subsection{Memory Layer}
\begin{figure*}
\begin{center}
\includegraphics[width=140mm,scale=1.5]{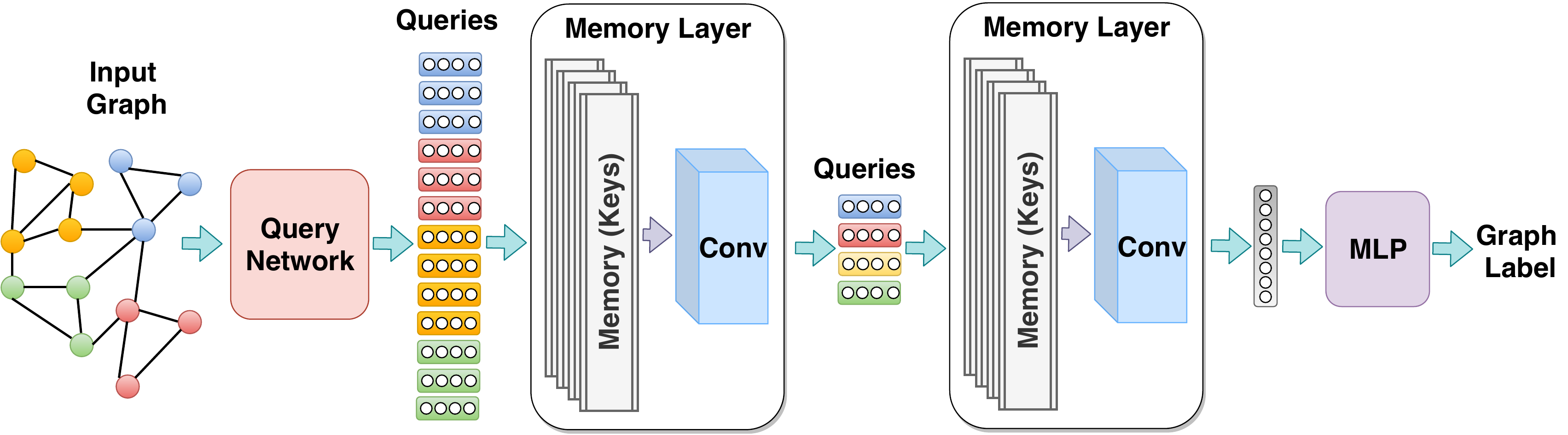}
\end{center}
   \caption{The proposed architecture for hierarchical graph representation learning using the proposed memory layer. The query network projects the initial node 
   features into a latent query space and each memory layer jointly coarsens the input queries and transforms them into a new query space.}
\label{fig:arch}
\end{figure*}

%------------------------------------------------------------------------------------------------------------------------------------------------------------------------------------------------------------------------
We define a memory layer $\mathcal{M}^{(l)}: \mathbb{R}^{n_l \times d_l}\longmapsto \mathbb{R}^{n_{l+1} 
\times d_{l+1}}$ in layer $l$ as a parametric function that takes in $n_l$ query vectors of size $d_l$ and generates $n_{l+1}$ query vectors 
of size $d_{l+1}$ such that $n_{l+1} < n_{l}$. The input and output queries represent the node representations of the input graph and the coarsened graph, respectively. The 
memory layer learns to jointly coarsen the input nodes, i.e., pooling, and transform their features, i.e., representation learning. As shown in Figure \ref{fig:arch}, a 
memory layer consists of arrays of memory keys, i.e., multi-head memory, and a convolutional layer. Assuming $|h|$ memory heads, a shared input query is 
compared against all the keys in each head resulting in $|h|$ attention matrices which are then aggregated into a single attention matrix using the convolution layer.

%------------------------------------------------------------------------------------------------------------------------------------------------------------------------------------------------------------------------
In a \textit{content addressable memory} \citep{graves_2014_arxiv, sukhbaatar_2015_nips, weston_2015_iclr}, the task of attending to memory, i.e., addressing scheme, is 
formulated as computing the similarity between memory keys to a given query $q$. Specifically, the attention weight of key $k_j$ for query $q$ is defined as 
$w_j=softmax(d(q, k_j))$ where $d$ is a similarity measure, typically Euclidean distance or cosine similarity \citep{rae_2016_nips}. The soft read operation on memory is 
defined as a weighted average over the memory keys: $r=\sum_j w_jk_j$.

%------------------------------------------------------------------------------------------------------------------------------------------------------------------------------------------------------------------------
In this work, we treat the input queries $\textbf{Q}^{(l)} \in \mathbb{R}^{n_l \times d_l}$ as the node representations of an input graph and treat the keys $\textbf{K}^{(l)} \in \mathbb{R}^{n_{l+1} \times d_{l}}$ as the cluster centroids of the 
queries. To satisfy this assumption, we impose a clustering-friendly distribution as the distance metric between keys and a query. Following \citep{xie_2016_icml, 
maaten_2008_jmlr}, we use the Student's t-distribution as a kernel to measure the normalized similarity between query $q_i$ and key ${k_j}$ as follows:

\begin{equation}
    C_{i,j} =\frac{{\left(1+{||q_i-{k_j}||}^2/\tau\right)}^{-\frac{\tau + 1}{2}}}
    {\sum\limits_{j'}{\left(1+{||q_i- {k_{j'}}||}^2/\tau\right)}^{-\frac{\tau + 1}{2}}} 
\end{equation}

where $C_{ij}$ is the normalized score between query $q_i$ and key ${k_j}$, i.e., probability of assigning node $i$ to cluster $j$ or attention score between query $q_i$ and memory key $k_j$, and $\tau$ is the degree of freedom of the Student's t-distribution, i.e., temperature. 
%------------------------------------------------------------------------------------------------------------------------------------------------------------------------------------------------------------------------
%------------------------------------------------------------------------------------------------------------------------------------------------------------------------------------------------------------------------
To increase the capacity, we model the memory keys as a multi-head array. Applying a shared input query against the memory keys produces a tensor of cluster 
assignments $\left[\textbf{C}^{(l)}_0 ...\textbf{C}^{(l)}_{|h|}\right] \in \mathbb{R}^{|h| \times n_{l+1} \times n_l} $ where $|h|$ denotes the number of heads. To aggregate the heads into a single assignment matrix, we treat the heads and the 
assignments matrices as depth, height, and width channels in standard convolution analogy and apply a convolution operator over them. Because there is no spatial 
structure, we use $[1 \times 1]$ convolution to aggregate the information across heads and therefore the convolution behaves as a weighted pooling that reduces the heads 
into a single matrix. The aggregated assignment matrix is computed as follows:

\begin{equation}
	\textbf{C}^{(l)}=\text{softmax}\left(\Gamma_{\phi}{\left({\operatorname*{\parallel}_{k=0}^{|h|} \textbf{C}^{(l)}_{k}}\right)}\right)
    \in\mathbb{R}^{n_{l} 
	\times n_{l+1}}
	\label{eq:c}
\end{equation}

where $\Gamma_{\phi}$ is a $[1 \times 1]$ convolutional operator parametrized by $\phi$, $||$ is the concatenation operator, and $\textbf{C}^{(l)}$ is the aggregated soft assignment matrix.

%------------------------------------------------------------------------------------------------------------------------------------------------------------------------------------------------------------------------
A memory read generates a value matrix $\textbf{V}^{(l)} \in \mathbb{R}^{n_{l+1} \times d_{l}}$ that represents the coarsened node representations in the same space as the input queries and is defined as 
the product of the soft assignment scores and the original queries as follows:

\begin{equation}
	\textbf{V}^{(l)}= \textbf{C}^{(l)\top}\textbf{Q}^{(l)} \in \mathbb{R}^{n_{l+1} \times d_{l}}
	\label{eq:6}
\end{equation}

The value matrix is fed to a single-layer feed-forward neural network to project the coarsened embeddings from $\mathbb{R}^{n_{l+1} \times d_{l}}$ into $\mathbb{R}^{n_{l+1} \times d_{l+1}}$ as follows: 

\begin{equation}
	\textbf{Q}^{(l+1)}=\sigma\left(\textbf{V}^{(l)}\textbf{W}\right) \in \mathbb{R}^{n_{l+1} \times d_{l+1}}
		\label{eq:7}
\end{equation}

where $\textbf{Q}^{(l+1)} \in \mathbb{R}^{n_{l+1} \times d_{l+1}}$is the output queries, $\textbf{W} \in  \mathbb{R}^{d_{l} \times d_{l+1}}$ is the network parameters, and $\sigma$ is the non-linearity implemented using LeakyReLU.

%------------------------------------------------------------------------------------------------------------------------------------------------------------------------------------------------------------------------
Thanks to these parametrized transformations, a memory layer can jointly learn the node representations and coarsens the graph end-to-end. The computed queries 
$\textbf{Q}^{(l+1)}$ are the input queries to the subsequent memory layer $\mathcal{M}^{(l+1)}$. For graph classification tasks, one can simply stack layers of memory up to the level where the 
input graph is coarsened into a single node representing the global graph representation and then feed it to a fully-connected layer to predict the graph class as 
follows:

\begin{equation}
	\mathcal{Y}=\text{softmax}\left(\text{MLP}\left(\mathcal{M}^{(l)}\left(\mathcal{M}^{(l-1)}\left(....\mathcal{M}^{(0)}\left(\textbf{Q}_0\right)\right)\right)\right)\right)
		\label{eq:7}
\end{equation}

where $\textbf{Q}_0=f_q(g)$ is the initial query representation\footnote{We use initial node representation and initial query representation interchangeably throughout the paper.} generated by the query network $f_q$ over graph $g$. We introduce two architectures based on the memory layer: GMN and MemGNN. These two architectures 
are different in the way that the query network is implemented. More specifically, GMN uses a feed-forward network for initializing the query: $f_q(g)=\text{FFN}_{\theta}(g)$, 
whereas MemGNN implements the query network as a message passing GNN: $f_q(g)=\text{GNN}_{\theta}(g)$. 
%------------------------------------------------------------------------------------------------------------------------------------------------------------------------------------------------------------------------
\subsection{GMN Architecture}
%------------------------------------------------------------------------------------------------------------------------------------------------------------------------------------------------------------------------
A GMN is a stack of memory layers on top of a query network $f_q(g)$ that generates the initial query representations without any message passing. Similar to set 
neural networks \citep{vinyals_2015_iclr} and transformers \citep{vaswani_2017_nips}, nodes in GMN are treated as a permutation-invariant set of representations. The 
query network projects the initial node features into a latent space that represents the initial query space. 

%------------------------------------------------------------------------------------------------------------------------------------------------------------------------------------------------------------------------
Assume a training set $\mathcal{D}=[g_1, g_2,...,g_N]$ of  $N$ graphs where each graph is represented as $g=(\textbf{A}, \textbf{X}, Y) $ and $\textbf{A}\in\lbrace 0, 1 \rbrace^ {n\times n}$ denotes the adjacency matrix, 
$\textbf{X}\in \mathbb{R}^{n \times d_{in}}$ is the initial node features, and $Y\in \mathbb{R}^{n}$ is the graph label. Considering that GMN treats a graph as a set of order-invariant nodes and does not use 
message passing, and also considering that the memory layers do not rely on connectivity information, the topological information of each node should be somehow 
encoded into its initial representation. To define the topological embedding, we use an instantiation of general graph diffusion matrix $\textbf{S} \in \mathbb{R}^{n \times n} $. More specifically, 
we use random walk with restart (RWR) \citep{pan_2004_kdd} to compute the topological embeddings and then sort them row-wise to force the embedding to be 
order-invariant. For further details please see section \ref{sec:topembedding}. Inspired by transformers \citep{vaswani_2017_nips}, we then fuse the topological 
embeddings with the initial node features into the initial query representations using a query network $f_q$ implemented as a two-layer feed-forward neural network:

\begin{equation}
	\textbf{Q}^{(0)}=\sigma \Big(\Big[\sigma(\textbf{SW}_0) \parallel \textbf{X}\Big]\textbf{W}_1\Big)
\end{equation}

where $\textbf{W}_0 \in  \mathbb{R}^{n \times d_{in}}$ and $\textbf{W}_1 \in  \mathbb{R}^{2d_{in} \times d_0}$ are the parameters of the query networks, $||$ is the concatenation operator, and $\sigma$ is the non-linearity implemented using 
LeakyReLU.

%------------------------------------------------------------------------------------------------------------------------------------------------------------------------------------------------------------------------
\subsubsection{Permutation Invariance}
\label{sec:perminv}
Considering the inherent permutation-invariant property of graph-structured data, a model designed to address graph classification tasks, should also enforce this 
property. This implies that the model should generate same outputs for isomorphic input graphs. We impose this on GMN architecture by sorting the topological 
embedding row-wise as a pre-processing step. 

\textbf{Proposition 1.} \textit{Given a sorting function o,} $\text{GMN}\left(o\left(\mathbf{S}\right), \mathbf{X}\right)$ \textit{is permutation-invariant.} \\

\textit{Proof.} Let $\textbf{P} \in \{ 0,1 \}^ { \{n \times n\} }$ be an arbitrary permutation matrix. For each node in graph $\textbf{G}$ with adjacency matrix $\textbf{A}$, the corresponding node in graph $\textbf{G}_P$ with 
permuted adjacency matrix $\textbf{PAP}^T$ has the permuted version of the topological embedding of the node in graph $\textbf{G}$. Sorting the embeddings cancels out the effect of 
permutation and makes the corresponding embeddings in graph $\textbf{G}$ and  $\textbf{G}_P$ identical.
%------------------------------------------------------------------------------------------------------------------------------------------------------------------------------------------------------------------------
\subsection{MemGNN Architecture}
%------------------------------------------------------------------------------------------------------------------------------------------------------------------------------------------------------------------------
Unlike the GMN architecture, the query network in MemGNN relies on message passing to compute the initial query $\textbf{Q}_0$ as follows:

\begin{equation}
	\textbf{Q}^{(0)}=G_{\theta}\left(\textbf{A}, \textbf{X}\right)
\end{equation}

where query network $G_\theta$ is an arbitrary parameterized message passing GNN \citep{gilmer_2017_icml, li_2015_iclr, kipf_2016_iclr, velickovic_2018_iclr}.

%------------------------------------------------------------------------------------------------------------------------------------------------------------------------------------------------------------------------
In our implementation, we use a modified variant of GAT \citep{velickovic_2018_iclr}. Specifically, we introduce an extension to the original GAT model called 
edge-based GAT (\textbf{e-GAT}) and use it as the query network. Unlike GAT, e-GAT learns attention weights not only from the neighbor nodes but also from the input 
edge features. This is especially important for data containing edge information (e.g., various bonds among atoms represented as edges in molecule datasets). In an 
e-GAT layer, attention score between two neighbor nodes is computed as follows:

\begin{equation}
	\alpha_{ij}=\frac{\text{exp}\left(\sigma\left(\textbf{W}\left[\textbf{W}_nh_i^{(l)} \parallel \textbf{W}_nh_j^{(l)} \parallel \textbf{W}_eh_{i\to j}^{(l)}  \right] \right) \right)}{\sum\limits_{k \in \mathcal{N}_i} \text{exp}\left(\sigma\left(\textbf{W}\left[\textbf{W}_nh_i^{(l)} \parallel \textbf{W}_nh_k^{(l)} \parallel \textbf{W}_eh_{i\to k}^{(l)}  \right] \right) \right)}
\end{equation}

where $h_i^{(l)}$and $h_{i \to j}^{(l)}$ denote the representation of node $i$ and the representation of the edge connecting node $i$ to its one-hop neighbor node $j$ in layer $l$, respectively. 
$\textbf{W}_n$ and $\textbf{W}_e$ are learnable node and edge weights and $\textbf{W}$ is the parameter of a single-layer feed-forward network that computes the attention score. $\sigma$ is the 
non-linearity implemented using LeakyReLU.
%------------------------------------------------------------------------------------------------------------------------------------------------------------------------------------------------------------------------
\subsection{Training}
%------------------------------------------------------------------------------------------------------------------------------------------------------------------------------------------------------------------------
We jointly train the model using two loss functions: a supervised loss and an unsupervised clustering loss. The supervised loss denoted as $\mathcal{L}_{sup}$ is defined as 
cross-entropy loss and root mean square error (RMSE) for graph classification and regression tasks, respectively. The unsupervised clustering loss is inspired by 
deep clustering methods \citep{razavi2019generating, xie_2016_icml, aljalbout2018clustering}. It encourages the model to learn clustering-friendly embeddings in the 
latent space by learning from high confidence assignments with the help of an auxiliary target distribution. The unsupervised loss is defined as the Kullback-Leibler 
(KL) divergence between the soft assignments $\textbf{C}^{(l)}$ and the auxiliary distribution $\textbf{P}^{(l)}$ as follows:

\begin{equation}
    \mathcal{L}_{KL}^{(l)} = \text{KL}\left(\textbf{P}^{(l)}||\textbf{C}^{(l)}\right) = \sum\limits_{i}\sum\limits_{j}
    P_{ij}^{(l)}log\frac{P_{ij}^{(l)}}{C_{ij}^{(l)}}
\end{equation}

For the target distributions $\textbf{P}^{(l)}$, we use the distribution proposed in \citep{xie_2016_icml} which normalizes the loss contributions and improves the cluster purity 
while emphasizing on the samples with higher confidence. This distribution is defined as follows:

\begin{equation}
    P_{ij}^{(l)} =\frac{\left(C_{ij}^{(l)}\right)^2/\sum\limits_{i}C_{ij}^{(l)}}
    {\sum\limits_{j'}\left(C_{ij'}^{(l)}\right)^2/\sum\limits_{i}C_{ij'}^{(l)}} 
\end{equation}
We define the total loss as follows where $L$ is the number of memory layers and $\lambda \in [0,1]$ is a scalar weight.

\begin{equation}
    \mathcal{L} = \frac{1}{N} \sum\limits_{n=1}^{N} 
    \left(\lambda\mathcal{L}_{sup} +(1-\lambda)\sum\limits_{l=1}^{L}\mathcal{L}_{KL}^{(l)}  \right)
    \label{eq:10}
\end{equation}

We initialize the model parameters, the keys, and the queries randomly and optimize them jointly with respect to $\mathcal{L}$ using mini-batch stochastic gradient descent. To 
stabilize the training, the gradients of $\mathcal{L}_{sup}$ are back-propagated batch-wise while the gradients of $\mathcal{L}_{KL}^{(l)}$ are applied epoch-wise by periodically switching $\lambda$ between 
$0$ and $1$. Updating the centroids, i.e., memory keys, with the same frequency as the network parameters can destabilize the training. To address this, we optimize all 
model parameters and the queries in each batch with respect to $\mathcal{L}_{sup}$ and in each epoch with respect to $\mathcal{L}_{KL}^{(l)}$. Memory keys, on the other hand, are only updated at 
the end of each epoch by the gradients of $\mathcal{L}_{KL}^{(l)}$. This technique has also been applied in \citep{hassani_2019_iccv, caron_2018_eccv} to avoid trivial solutions.
%------------------------------------------------------------------------------------------------------------------------------------------------------------------------------------------------------------------------
\section{Experiments}
\subsection{Datasets}
%------------------------------------------------------------------------------------------------------------------------------------------------------------------------------------------------------------------------
We use nine benchmarks including seven graph classification and two graph regression datasets to evaluate the proposed method. These datasets are commonly 
used in both graph kernel \citep{borgwardt_2005_icdm, yanardag_2015_kdd, shervashidze_2009_ais, ying_2018_nips,shervashidze_2011_jmlr,kriege_2016_nips} and 
GNN \citep{cangea_2018_nips, ying_2018_nips, lee_2019_icml, gao_2019_icml} literature. The summary of the datasets is as follows where the first two benchmarks are 
regression tasks and the rest are classification tasks.

%------------------------------------------------------------------------------------------------------------------------------------------------------------------------------------------------------------------------
\textbf{ESOL} \citep{delaney2004esol} contains water solubility data for compounds. \\
\textbf{Lipophilicity} \citep{gaulton2016chembl} contains experimental results of octanol/water distribution of compounds.  \\
\textbf{Bace} \citep{subramanian2016computational} provides quantitative binding results for a set of inhibitors of human $\beta$-secretase 1 (BACE-1).\\
\textbf{DD} \citep{dobson_2003_jmb} is used to distinguish enzyme structures from non-enzymes. \\
\textbf{Enzymes} \citep{schomburg2004brenda} is for predicting functional classes of enzymes. \\
\textbf{Proteins} \citep{dobson_2003_jmb}  is used to predict the protein function from structure. \\
\textbf{Collab} \citep{yanardag_2015_kdd} is for predicting the field of a researcher given one's ego-collaboration graph. \\
\textbf{Reddit-Binary} \citep{yanardag_2015_kdd} is for predicting the type of community given a graph of online discussion threads.\\ 
\textbf{Tox21} \citep{challenge2014tox21} is for predicting toxicity on 12 different targets.
 	
For more information about the detailed statistics of the datasets refer to Appendix \ref{sec:datastat}.
%------------------------------------------------------------------------------------------------------------------------------------------------------------------------------------------------------------------------
\subsection{Graph Classification Results}
\label{ressubsec}
To evaluate the performance of our models on DD, Enzymes, Proteins, Collab, and Reddit-Binary datasets, we follow the experimental protocol in 
\citep{ying_2018_nips} and perform 10-fold cross-validation and report the mean accuracy over all folds. We also report the performance of four graph kernel methods 
including Graphlet \citep{shervashidze_2009_ais}, shortest path \citep{borgwardt_2005_icdm}, Weisfeiler-Lehman (WL) \citep{shervashidze_2011_jmlr}, and WL Optimal 
Assignment \citep{kriege_2016_nips}, and ten GNN models. 

%------------------------------------------------------------------------------------------------------------------------------------------------------------------------------------------------------------------------
The results shown in Table \ref{tab:pool} suggest that: (\textit{i}) our models achieve state-of-the-art results w.r.t. GNN models and significantly improve the performance 
on Enzymes, Proteins, DD, Collab, and Reddit-Binary datasets by absolute margins of 14.49\%, 6.0\%,  3.76\%, 2.62\%, and 8.98\% accuracy, respectively, (\textit{ii}) our 
models outperform graph kernels on all datasets except Collab where our models are competitive with the best kernel, i.e., absolute margin of 0.56\%, (\textit{iii}) both 
proposed models achieve better performance or are competitive compared to the baseline GNNs, (\textit{iv}) GMN achieves better results compared to MemGNN which 
suggests that replacing local adjacency information with global topological embeddings provides the model with more useful information, and (\textit{v}) On Collab, our 
models are outperformed by a variant of DiffPpool (i.e., diffpool-det) \citep{ying_2018_nips} and WL Optimal Assignment \citep{kriege_2016_nips}. The former is a GNN 
augmented with deterministic clustering algorithm\footnote{In diffpool-det assignment matrices are generated using a deterministic graph clustering algorithm.}, 
whereas the latter is a graph kernel method. We speculate this is because of the high edge-to-node ratio in this dataset and the augmentations used in these two 
methods help them with extracting near-optimal cliques.
%------------------------------------------------------------------------------------------------------------------------------------------------------------------------------------------------------------------------
\begin{table}[t]
\begin{center}
\addtolength{\tabcolsep}{-1.8pt} 
\caption{Mean validation accuracy over 10-folds.}
\vspace{-0.2cm}
\label{tab:pool}
\centering
\begin{tabular}{llccccc}
\hline
\multicolumn{2}{c}{\textbf{Method}} &\multicolumn{5}{c}{\textbf{Dataset}} \\
\cline{3-7}
& & Enzymes & Proteins	& DD & Collab & Reddit-B \\
\hline
\multirow{4}{*}{\begin{turn}{90}Kernel\end{turn}}
& Graphlet \citep{shervashidze_2009_ais} & 41.03 & 72.91 & 64.66  & 64.66 & 78.04\\
& ShortestPath \citep{borgwardt_2005_icdm}    & 42.32 & 76.43 & 78.86 & 59.10 & 64.11\\
& WL \citep{shervashidze_2011_jmlr}  & 53.43  & 73.76 & 74.02 & 78.61 & 68.20\\
& WL Optimal \citep{kriege_2016_nips} & 60.13 & 75.26 & 79.04 & \textbf{80.74} & 89.30\\
\hline 
\multirow{12}{*}{\begin{turn}{90}GNN\end{turn}}
& PatchySan \citep{niepert_2016_icml} & $-$ & 75.00 &  76.27 & 72.60 & 86.30 \\
& GraphSage \citep{hamilton_2017_nips}  & 54.25 & 70.48 & 75.42 & 68.25 & $-$\\
& ECC \citep{simonovsky_2017_cvpr} & 53.50 & 72.65 & 74.10 & 67.79 & $-$\\
& Set2Set \citep{vinyals_2015_iclr} & 60.15 & 74.29 & 78.12 & 71.75 & $-$\\
& SortPool \citep{morris_2019_aaai} & 57.12 & 75.54 & 79.37 & 73.76 & $-$\\
& DiffPool \citep{ying_2018_nips} &   60.53 & 76.25 & 80.64 & 75.48 & 85.95\\
& CliquePool \citep{Luzhnica_2019_iclr} & 60.71 & 72.59 & 77.33 & 74.50 & $-$\\
& Sparse HGC \citep{cangea_2018_nips} & 64.17 & 75.46 & 78.59 &  75.46 & 79.20\\
% Mincut Pool \citep{Bianchi_2019_arxiv} & $-$ & 76.5 & 80.3 & \textbf{83.4}\\
& TopKPool \citep{gao_2019_icml} & $-$ & 77.68 &  82.43 & 77.56 & 74.70\\
& SAGPool \citep{lee_2019_icml} & $-$ & 71.86 & 76.45 & $-$ & 73.90 \\
\cline{2-7}
& GMN (ours) &  \textbf{78.66}  & \textbf{82.25} & \textbf{84.40} &  80.18 & \textbf{95.28}\\
& MemGNN (ours) & 75.50  & 81.35 & 82.92 &  77.0 & 85.55\\
\hline
\end{tabular}
\end{center}
\end{table}
%------------------------------------------------------------------------------------------------------------------------------------------------------------------------------------------------------------------------
\begin{table}[t]
\begin{center}
  \caption{AUC-ROC on BACE and Tox21 datasets.}
  \vspace{-0.2cm}
  \label{tab:bacetox}
  \centering
  \begin{tabular}{lcc|cc}
    \hline
  \bf{Method} &  \multicolumn{4}{c}{\bf{Dataset}}\\
  \cline{2-5}
    & \multicolumn{2}{c}{BACE} & \multicolumn{2}{c}{Tox21} \\
    \cline{2-3}
    \cline{4-5}
    ~ & \multicolumn{1}{c}{validation} & \multicolumn{1}{c}{test} & \multicolumn{1}{c}{validation} & \multicolumn{1}{c}{test} \\
    \hline
    Logistic Regression    & 0.719 $\pm$ 0.003    & 0.781 $\pm$ 0.010  & 0.772 $\pm$ 0.011   & 0.794 $\pm$ 0.015 \\
    KernelSVM & 0.739 $\pm$ 0.000  & 0.862 $\pm$ 0.000  & 0.818 $\pm$ 0.010  & 0.822 $\pm$ 0.006 \\
    XGBoost & 0.756 $\pm$ 0.000  & 0.850 $\pm$ 0.008 & 0.775 $\pm$ 0.018  & 0.794 $\pm$ 0.014 \\
    Random Forest & 0.728 $\pm$ 0.004 & 0.867 $\pm$ 0.008  & 0.763 $\pm$ 0.002   & 0.769 $\pm$ 0.015 \\
    IRV & 0.715 $\pm$ 0.001 & 0.838 $\pm$ 0.000 & 0.807 $\pm$ 0.006 & 0.799 $\pm$ 0.006 \\
    Multitask & 0.696 $\pm$ 0.037 & 0.824 $\pm$ 0.0006   & 0.795 $\pm$ 0.017 & 0.803 $\pm$ 0.012 \\
    Bypass & 0.745 $\pm$ 0.017 & 0.829 $\pm$ 0.006  & 0.800 $\pm$ 0.008 & 0.810 $\pm$ 0.013 \\
    GCN    & 0.627 $\pm$ 0.015   & 0.783 $\pm$ 0.014  & 0.825 $\pm$ 0.013 & \bf{0.829 $\pm$ 0.006} \\
    Weave    & 0.638 $\pm$ 0.014   & 0.806 $\pm$ 0.002 & 0.828 $\pm$ 0.008 & 0.820 $\pm$ 0.010 \\
    \hline
    MemGNN (ours)  & \bf{0.859 $\pm$ 0.000}   & \bf{0.907 $\pm$ 0.000} & \bf{0.862 $\pm$ 0.009} & 0.828  $\pm$ 0.004\\
    \hline
  \end{tabular}
\end{center}
\end{table}
%------------------------------------------------------------------------------------------------------------------------------------------------------------------------------------------------------------------------
To evaluate the performance on the BACE and Tox21 datasets, we follow the evaluation protocol in \citep{wu2018moleculenet} and report the area under the curve 
receiver operating characteristics (AUC-ROC) measure. Considering that the BACE and Tox21 datasets contain initial edge features, we train the MemGNN model 
and compare its performance to the baselines reported in \citep{wu2018moleculenet}. The results shown in Table \ref{tab:bacetox} suggest that our model achieves 
state-of-the-art results by absolute margin of 4.0 AUC-ROC on the BACE benchmark and is competitive with the state-of-the-art GCN model on the Tox21 dataset, 
i.e., absolute margin of 0.001.
%------------------------------------------------------------------------------------------------------------------------------------------------------------------------------------------------------------------------
\subsection{Graph Regression Results}
\label{ressubsec}
For the ESOL and Lipophilicity datasets, we follow the evaluation protocol in \citep{wu2018moleculenet} and report their RMSEs. Considering that these datasets 
contain initial edge features (refer to Appendix \ref{sec:datastat} for further details), we train the MemGNN model and compare the results to the baseline models 
reported in \citep{wu2018moleculenet} including graph based methods such as GCN, MPNN, directed acyclic graph (DAG) model, and Weave as well as other 
conventional methods such as kernel ridge regression (KRR) and influence relevance voting (IRV). Results shown in Table \ref{tab:esollip} suggest that our 
MemGNN model achieves state-of-the-art results by absolute margin of 0.07 and 0.1 RMSE on ESOL and Lipophilicity benchmarks, respectively. For further details 
on regression datasets and baselines please refer to \citep{wu2018moleculenet}.
%------------------------------------------------------------------------------------------------------------------------------------------------------------------------------------------------------------------------
\begin{table}[t]
\begin{center}
  \caption{RMSE on ESOL and Lipophilicity datasets.}
  \vspace{-0.2cm}
  \label{tab:esollip}
  \centering
  \begin{tabular}{lcc|cc}
    \hline
  \bf{Method} &  \multicolumn{4}{c}{\bf{Dataset}}\\
  \cline{2-5}
    & \multicolumn{2}{c}{ESOL} & \multicolumn{2}{c}{Lipophilicity} \\
    \cline{2-3}
    \cline{4-5}
    ~ & \multicolumn{1}{c}{validation} & \multicolumn{1}{c}{test} & \multicolumn{1}{c}{validation} & \multicolumn{1}{c}{test} \\
    \hline
    Multitask & 1.17 $\pm$ 0.13 & 1.12 $\pm$ 0.19	&  0.852 $\pm$ 0.048 & 0.859 $\pm$ 0.013    \\
    Random Forest & 1.16 $\pm$ 0.15  & 1.07 $\pm$ 0.19 &  0.835 $\pm$ 0.036 &  0.876 $\pm$ 0.040   \\
    XGBoost  & 1.05 $\pm$ 0.10 & 0.99 $\pm$ 0.14 & 0.783 $\pm$ 0.021 &  0.799 $\pm$ 0.054  \\
    GCN     & 1.05 $\pm$ 0.15 & 0.97 $\pm$ 0.01 & 0.678 $\pm$ 0.040 & 0.655 $\pm$ 0.036 \\
    MPNN  & 0.55 $\pm$ 0.02 & 0.58 $\pm$ 0.03 & 0.757 $\pm$ 0.030 & 0.715 $\pm$ 0.035 \\
    KRR    & 1.65 $\pm$ 0.19 & 1.53 $\pm$ 0.06 & 0.889 $\pm$ 0.009 & 0.899 $\pm$ 0.043  \\
    DAG   & 0.74 $\pm$  0.04 & 0.82 $\pm$ 0.08 & 0.857 $\pm$ 0.050 & 0.835 $\pm$ 0.039 \\
    Weave & 0.57 $\pm$ 0.04 & 0.61 $\pm$ 0.07 & 0.734 $\pm$ 0.011 & 0.715 $\pm$ 0.035 \\
    \hline
    MemGNN (ours)  & \bf{0.53 $\pm$ 0.03} & \bf{0.54 $\pm$ 0.01} & \bf{0.555 $\pm$ 0.039} & \bf{0.556 $\pm$ 0.023} \\
    \hline
  \end{tabular}
  \end{center}
  \end{table}
%------------------------------------------------------------------------------------------------------------------------------------------------------------------------------------------------------------------------
\subsection{Ablation Study}
\subsubsection{Effect of edge features \label{sec:edgeeffect}}
To investigate the effect of the proposed e-GAT model, we train the MemGNN model using both GAT and e-GAT layers as the query network. Considering that the ESOL, Lipophilicity, and BACE datasets contain edge features, we use them as the benchmarks. Since nodes have richer features compared to edges, we set the node 
and edge feature dimensions to 16 and 4, respectively. The performance of the two layers on the ESOL dataset shown in Appendix\ref{sec:gatcompare} suggesting 
that e-GAT achieves better results on the validation set in each epoch compared to the standard GAT model. We observed the same effect on Lipophilicity and BACE 
datasets.
%------------------------------------------------------------------------------------------------------------------------------------------------------------------------------------------------------------------------
\subsubsection{Effect of Topological Embedding}
To investigate the effect of topological embeddings on the GMN model, we evaluated three initial topological features including adjacency matrix, normalized 
adjacency matrix, and RWR. For further details on RWR, see section \ref{sec:rwr}. The results suggest that using the RWR as the initial positional embedding achieves 
the best performance. For instance, 10-fold cross validation accuracy of a GMN model trained on Enzymes dataset with adjacency matrix, normalized adjacency 
matrix, and RWR are 78.66\%, 77.16\%, and 77.33\%, respectively. Furthermore, sorting the topological embeddings to guarantee invariance to permutations 
improves the performance. For example, it increases the accuracy on the DD dataset from 82.24\% to 84.40\%.
%------------------------------------------------------------------------------------------------------------------------------------------------------------------------------------------------------------------------
\subsubsection{Down-sampling Neighbors with Random Walks}
We investigate two methods to down-sample the neighbors in dense datasets such as Collab (i.e., average of 66 neighbors per node) to enhance the memory and 
computation. The first method randomly selects 10\% of the edges whereas the second method ranks the neighbors based on their RWR scores with respect to the 
center node and then keeps the top 10\% of the edges. We trained the MemGNN model on Collab using both sampling methods which resulted in 73.9\% and 
73.1\% 10-fold cross validation accuracy for random and RWR-based sampling methods respectively, suggesting that random sampling performs slightly better 
than RWR based sampling.
%------------------------------------------------------------------------------------------------------------------------------------------------------------------------------------------------------------------------
\subsubsection{Effect of Number of Keys and Heads}
We speculate that although keys represent the clusters, the number of keys is not necessarily proportional to the number of the nodes in the input graphs. In fact, 
datasets with smaller graphs might have more meaningful clusters to capture. For example, molecules are comprised of many functional groups and yet the average 
number of nodes in the ESOL dataset is 13.3. Moreover, our experiments show that for Enzymes with average number of 32.69 nodes, the best performance is 
achieved with 10 keys whereas for the ESOL dataset 64 keys results in the best performance. In ESOL 8, 64, and 160 keys result in RMSE of 0.56, 0.52, and 0.54, 
respectively. We also observed that with a fixed number of parameters, increasing the number of memory heads improves the performance. For instance, when the 
model is trained on ESOL with 160 keys and 1 head, it achieves RMSE of 0.54, whereas when trained with 32 keys and 5 heads, the same model achieves RMSE of 
0.53.
%------------------------------------------------------------------------------------------------------------------------------------------------------------------------------------------------------------------------
\begin{figure*}[t]
\begin{subfigure}[t]{.5\textwidth}
\begin{center}
\includegraphics[width=35mm,scale=1]{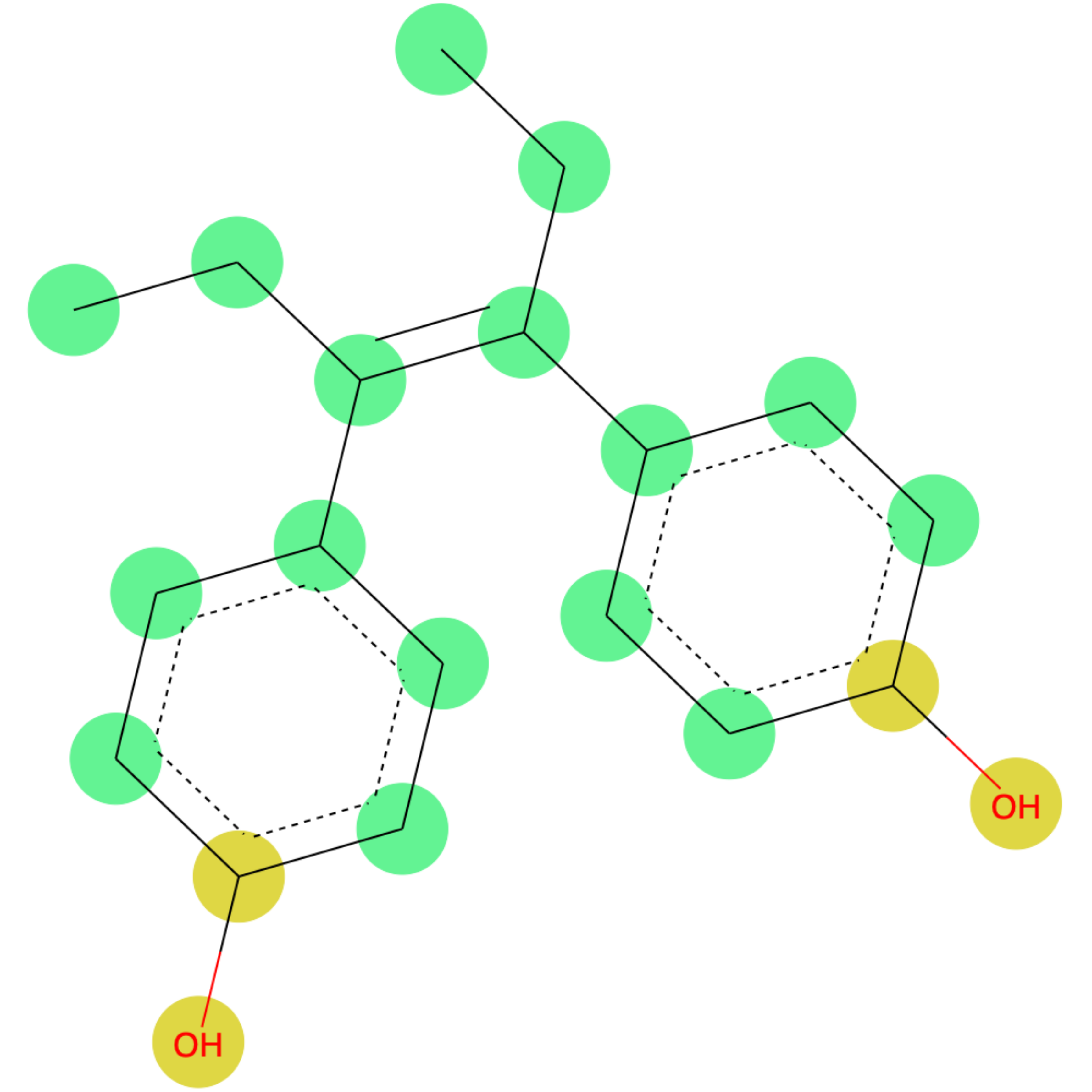}
\caption{}
\label{fig:mol1}
\end{center}
\end{subfigure}
\begin{subfigure}[t]{.5\textwidth}
\begin{center}
\includegraphics[width=45mm,scale=.4]{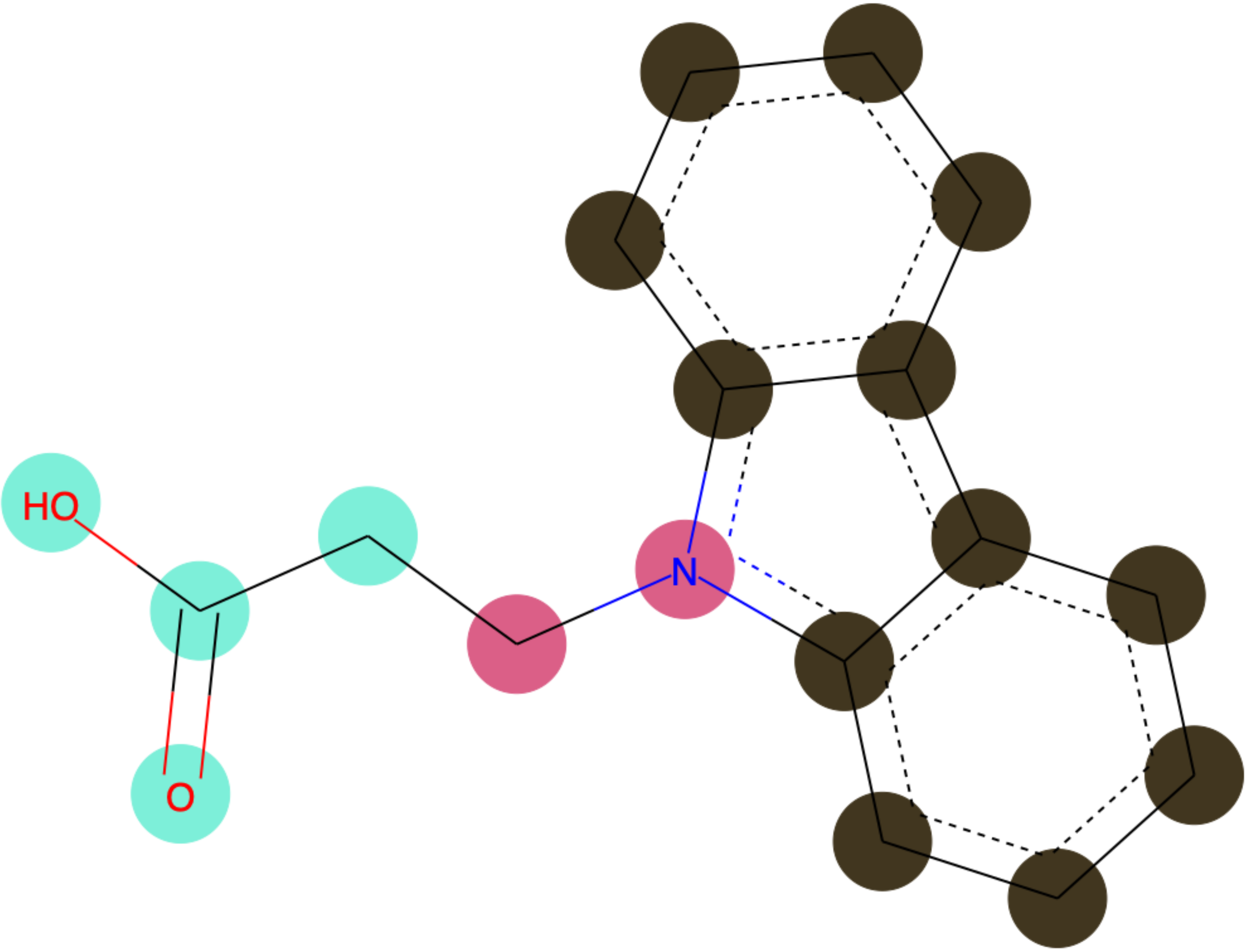}
\caption{}
\label{fig:mol2}
\end{center}
\end{subfigure}
\caption{Visualization of the learned clusters of two molecule instances from (a) ESOL and (b) Lipophilicity datasets. The visualizations show that the learned 
clusters correspond to known chemical groups. Note that a node without label represents a carbon atom. For more visualizations and discussion see section 
\ref{sec:vis2} }
\label{fig:mol}
\end{figure*}
%------------------------------------------------------------------------------------------------------------------------------------------------------------------------------------------------------------------------
\subsubsection{What Do the Keys Represent?}
Intuitively, the memory keys represent the cluster centroids and enhance the model performance by capturing meaningful structures. To investigate this, we used 
the learned keys to interpret the knowledge learned by models through visualizations. Figure \ref{fig:mol} visualizes the learned clusters over atoms (i.e., atoms with 
the same color are within the same cluster) indicating that the clusters mainly consist of meaningful chemical substructures such as a carbon chain and a Hydroxyl 
group (OH) (i.e., Figure \ref{fig:mol1}), as well as a Carboxyl group (COOH) and a benzene ring (i.e., Figure \ref{fig:mol2}). From a chemical perspective, Hydroxyl 
and Carboxyl groups, and carbon chains have a significant impact on the solubility of the molecule in water or lipid. This confirms that the network has learned 
chemical features that are essential for determining the molecule solubility. It is noteworthy that we tried initializing the memory keys using K-Means algorithm 
over the initial node representations to warm-start them but did not observe any significant improvement over the randomly selected keys.
%------------------------------------------------------------------------------------------------------------------------------------------------------------------------------------------------------------------------
\section{Conclusion}
We proposed an efficient memory layer and two models for hierarchical graph representation learning. We evaluated the proposed models on nine graph 
classification and regression tasks and achieved state-of-the-art results on eight of them. We also experimentally showed that the learned representations can capture 
well-known chemical features of the molecules. Furthermore, we showed that concatenating node features with topological embeddings and passing them through 
a few memory layers achieves notable results without using message passing. Moreover, we showed that defining the topological embeddings using graph diffusion 
achieves best performance. Finally, we showed that although connectivity information is not explicitly imposed on the model, the memory layer can process node 
representations and properly cluster and aggregate the learned representations.

%------------------------------------------------------------------------------------------------------------------------------------------------------------------------------------------------------------------------
\textbf{Limitations}: In section \ref{ressubsec}, we discussed that a graph kernel and a GNN augmented with deterministic clustering achieve better performance compared 
to our models on the Collab dataset. Analyzing samples in this dataset suggests that in graphs with dense communities, such as cliques, our model struggles to 
properly detect the dense sub-graphs.

%------------------------------------------------------------------------------------------------------------------------------------------------------------------------------------------------------------------------
\textbf{Future Directions}: We plan to extend our models to also perform node classification by attending to the node representations and centroids of the clusters from 
different layers of hierarchy that the nodes belongs to. Moreover, we are planning to evaluate other graph diffusion such as heat kernel, to initialize the topological 
embeddings. We are also planing to investigate the representation learning capabilities of the proposed models in self-supervised setting.

\bibliography{iclr2020_conference}
\bibliographystyle{iclr2020_conference}

\clearpage 
\appendix
\section{Appendix}
\subsection{Implementation Details}
\label{sec:impdet}
%------------------------------------------------------------------------------------------------------------------------------------------------------------------------------------------------------------------------
We implemented the model with PyTorch \citep{paszke2017automatic} and optimized it using Adam \citep{Kingma_2014_ICLR} optimizer. We trained the model for a 
maximum number of 2000 epochs and decayed the learning rate every 500 epochs by 0.5. The model uses batch-normalization \citep{Ioffe_2015_ICML}, 
skip-connections, LeakyRelu activation functions, and dropout \citep{Srivastava_2014_JMLR} for regularization. We also set the temperature in Student’s t-distribution 
to 1.0 and the restart probability in RWR to 0.1. We decided the hidden dimension and the number of model parameters using random hyper-parameter search 
strategy. The best performing hyper-parameters for the datasets are shown in Table \ref{tab:hyper}.
%------------------------------------------------------------------------------------------------------------------------------------------------------------------------------------------------------------------------
\subsection{Dataset Statistics \label{sec:datastat}}
Table \ref{tab:datastat} summarizes the statistics of the datasets used for graph classification and regression tasks.
%------------------------------------------------------------------------------------------------------------------------------------------------------------------------------------------------------------------------
\subsection{effect of e-gat \label{sec:gatcompare}}
\begin{figure}
\centering
\begin{subfigure}{.5\textwidth}
  \centering
  \includegraphics[width=1\linewidth]{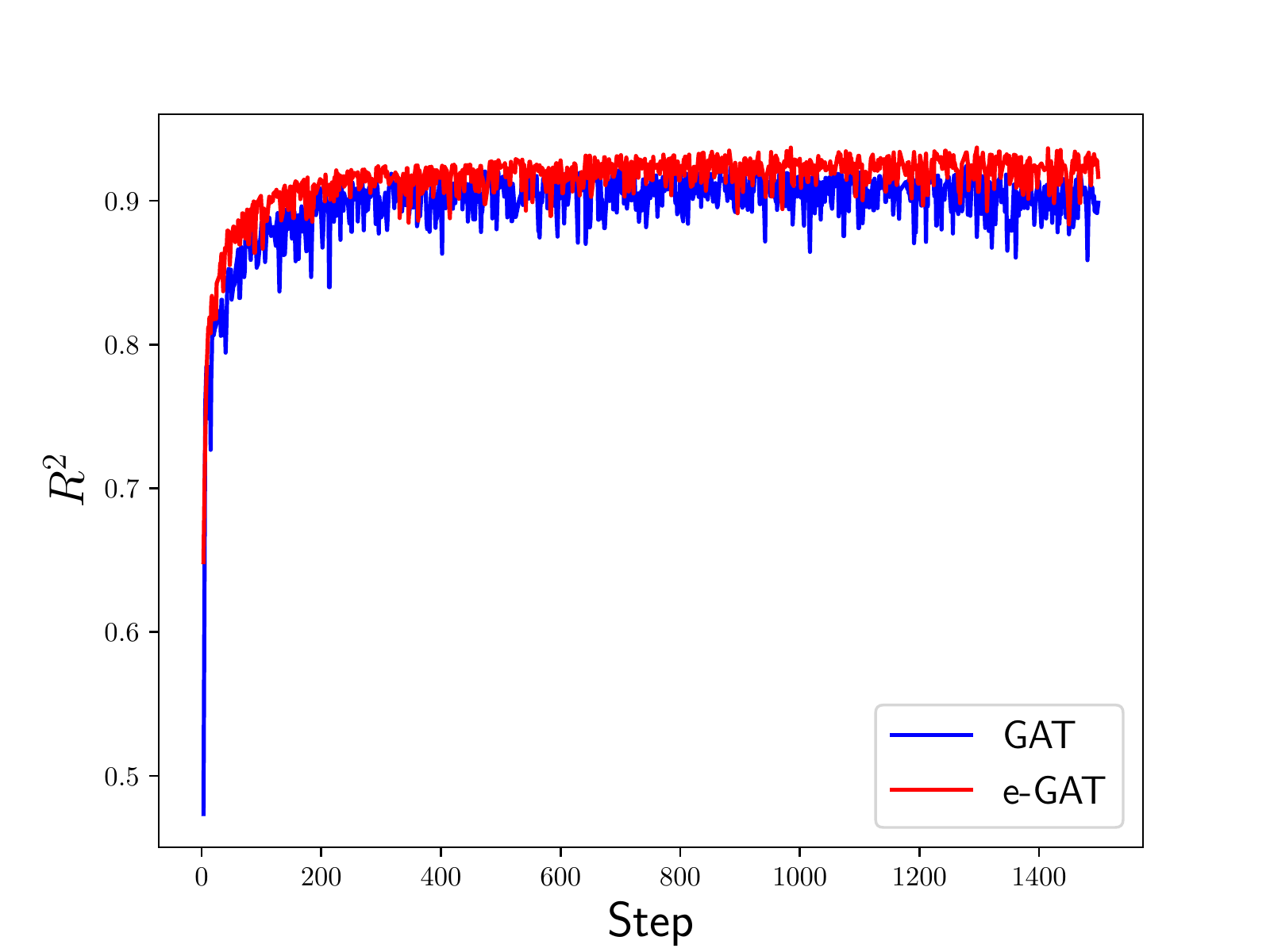}
  \caption{}
  \label{fig:rmse}
\end{subfigure}%
\begin{subfigure}{.5\textwidth}
  \centering
  \includegraphics[width=1\linewidth]{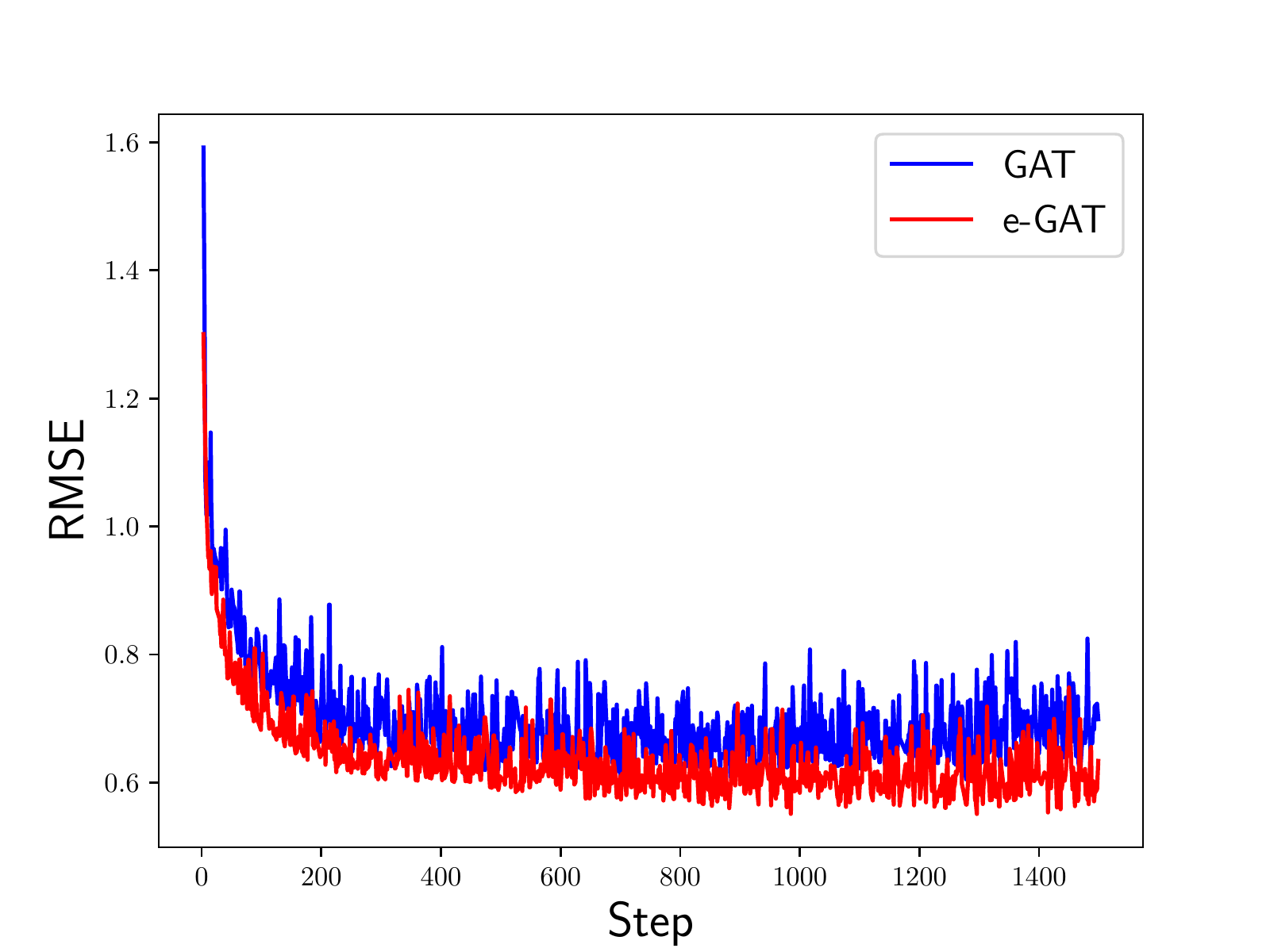}
  \caption{}
  \label{fig:r2}
\end{subfigure}
\caption{Validation (a) R$^2$ score, and (b) RMSE achieved by MemGNN model on ESOL with GAT and e-GAT based query networks.}
\label{fig:egat_comp}
\end{figure}

In section \ref{sec:edgeeffect}, we introduced e-GAT. Figures \ref{fig:rmse} and  \ref{fig:r2} illustrate the RMSE and R$^2$ score on the validation set of the ESOL dataset 
achieved by a MemGNN model using both GAT and e-GAT as the query network, respectively. As shown, e-GAT performs better compared to GAT on both metrics.
%------------------------------------------------------------------------------------------------------------------------------------------------------------------------------------------------------------------------
\subsection{Random Walk with Restart \label{sec:rwr}}
\label{sec:topembedding}
Suppose an agent randomly traverses a graph starting from node $i$ and iteratively walks towards its neighbors with a probability proportional to the edge weight 
that connects them. The agent also can randomly restart the traverse with probability $p$. Eventually, the agent will stop at node $j$ with a probability called relevance 
score of node $j$ with respect to node $i$ \citep{tong2006fast}. The relevance score of node $i$ with every other node of the graph is defined as follows: 
%------------------------------------------------------------------------------------------------------------------------------------------------------------------------------------------------------------------------
\begin{equation}
    \label{eq:12}
    \vec{t_i} = p\tilde{\textbf{A}}\vec{t_i} + (1-p)\vec{e_i}= (1-p)(I-p \tilde{\textbf{A}} )^{-1}\vec{e_i}
\end{equation}
%------------------------------------------------------------------------------------------------------------------------------------------------------------------------------------------------------------------------
where $\vec{t_i}$ is the RWR score corresponding to node $i$, $p$ is the restart probability, $\tilde{\textbf{A}}$ is the normalized adjacency matrix, and $\vec{e_i}$ is one-hot vector representation of 
node $i$. 

%------------------------------------------------------------------------------------------------------------------------------------------------------------------------------------------------------------------------
Note that the restart probability defines how far the agent can walk from the source node and therefore $\vec{t_i}$ represents the trad-off between local and global 
information around node $i$.
%------------------------------------------------------------------------------------------------------------------------------------------------------------------------------------------------------------------------
\subsection{Learned Clusters \label{sec:vis2}}
\begin{figure*}
    \centering
    \begin{subfigure}[b]{0.475\textwidth}
        \centering
        \includegraphics[width=\textwidth]{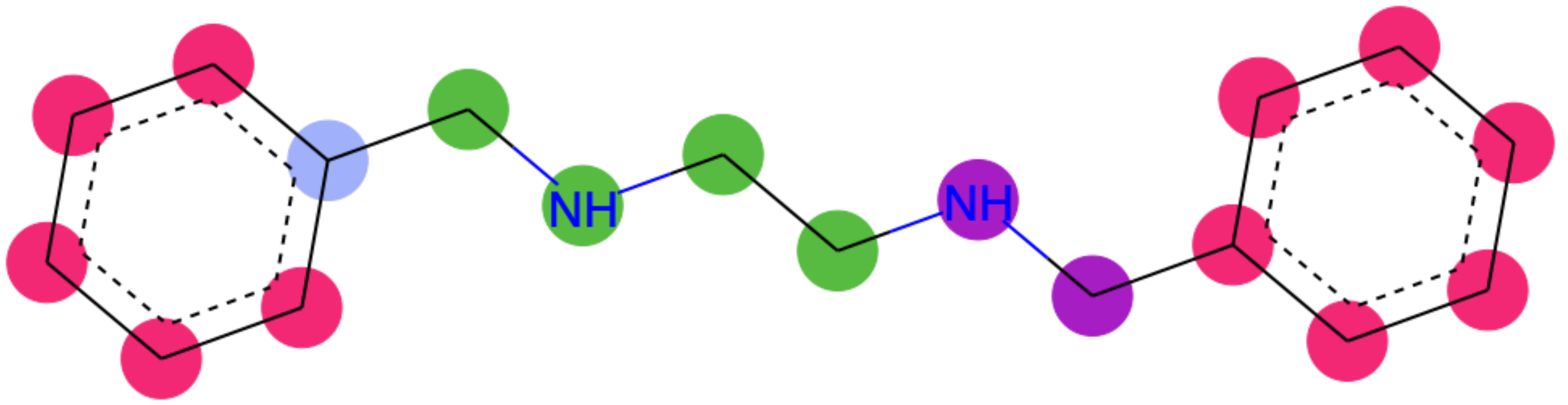}
        \caption{}   
        \label{fig:c_f_comp1}
    \end{subfigure}
    \hfill
    \begin{subfigure}[b]{0.475\textwidth}  
        \centering 
        \includegraphics[width=\textwidth]{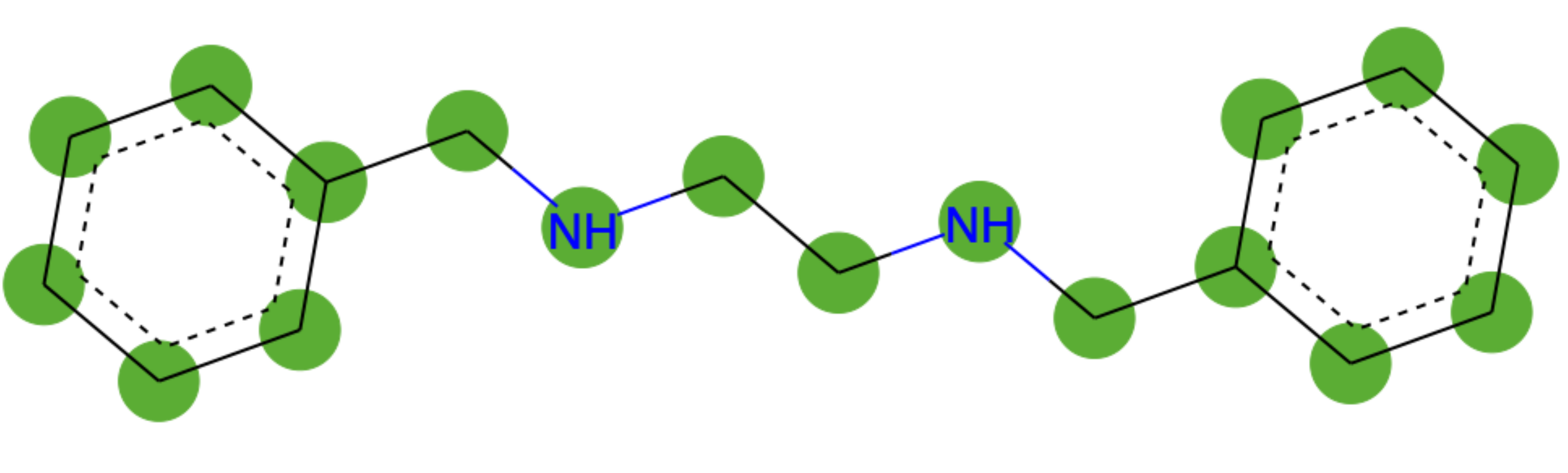}
        \caption{}    
        \label{fig:c_f_comp2}
    \end{subfigure}
    \vskip\baselineskip
    \begin{subfigure}[b]{0.475\textwidth}   
        \centering 
        \includegraphics[width=\textwidth]{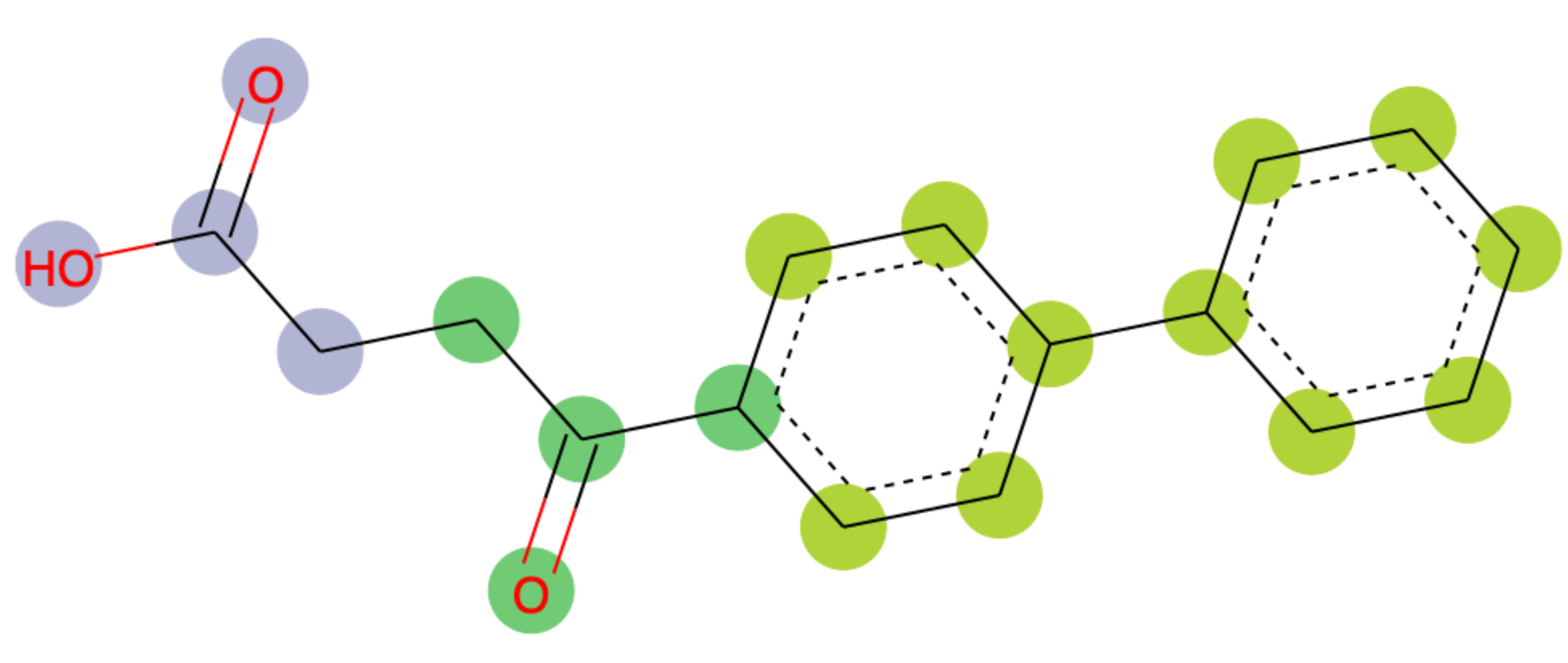}
        \caption{}    
        \label{fig:c_f_comp3}
    \end{subfigure}
    \quad
    \begin{subfigure}[b]{0.475\textwidth}   
        \centering 
        \includegraphics[width=\textwidth]{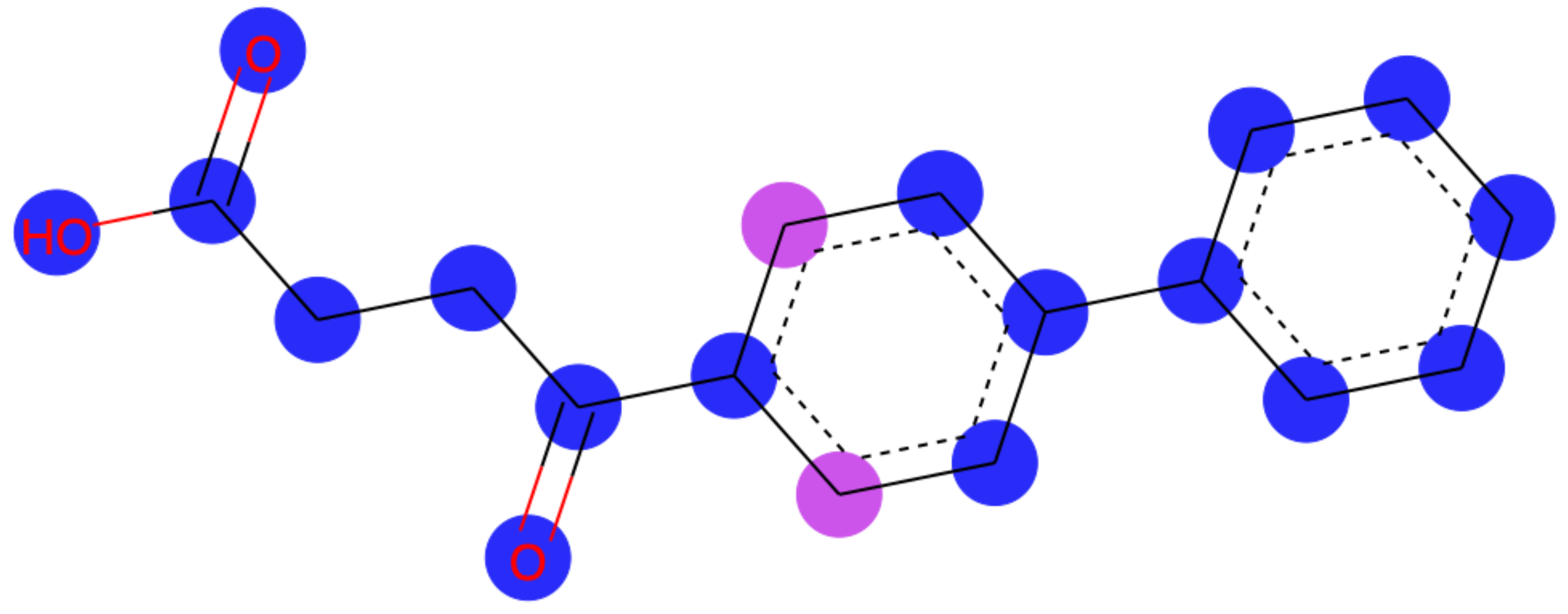}
        \caption{}    
        \label{fig:c_f_comp4}
    \end{subfigure}
    \caption{Figures (b) and (d) show computed clusters without using unsupervised clustering loss, whereas Figures (a) and (c) show the clusters 
    learned using the unsupervised clustering loss. The visualizations suggest that the unsupervised loss helps the model in learning distinct 
    and meaningful clusters.} 
    \label{fig:C_F_comp}
\end{figure*}

\begin{figure*}
    \centering
    \begin{subfigure}[b]{0.475\textwidth}
        \centering
        \includegraphics[width=\textwidth]{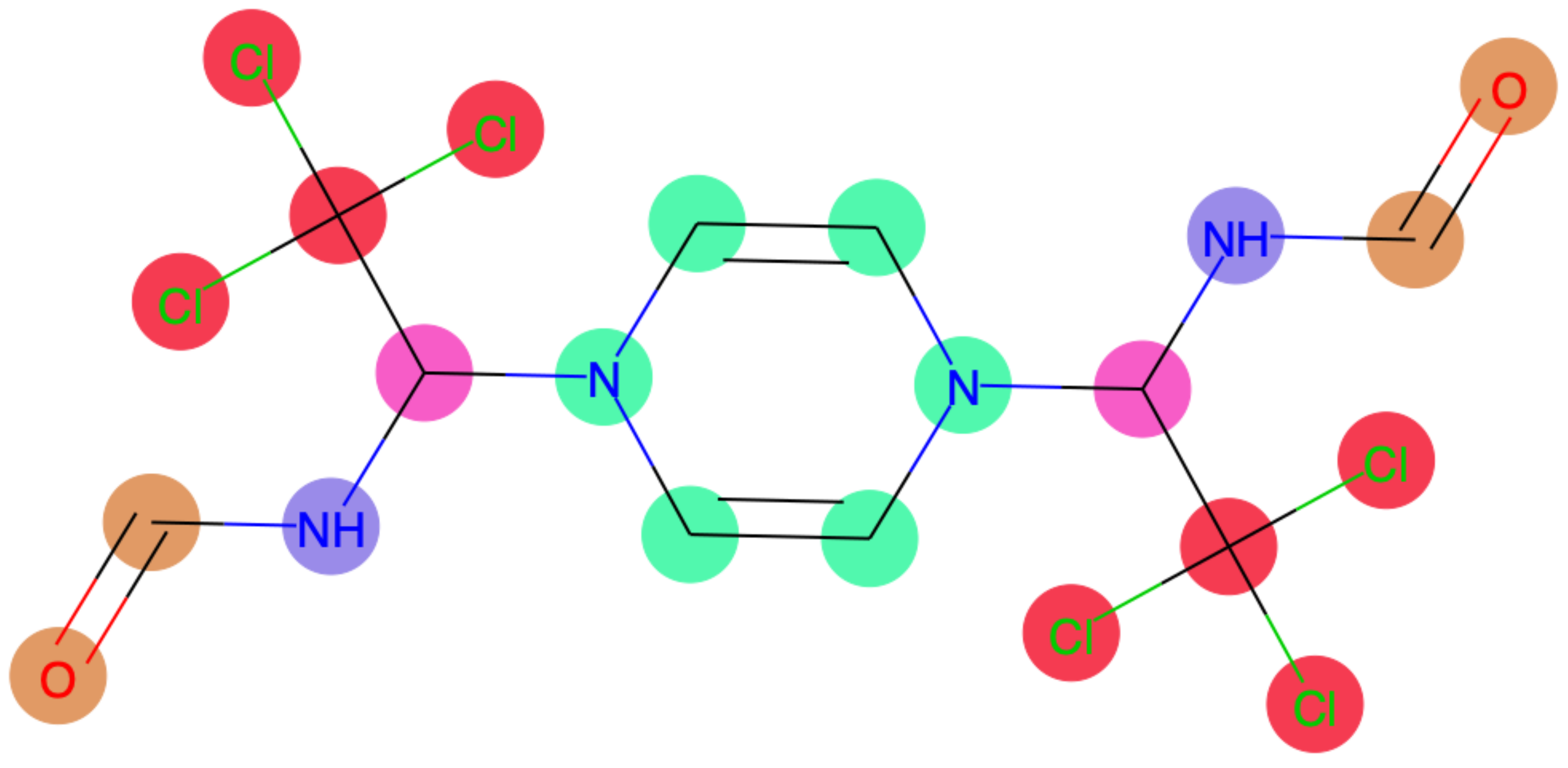}
        \caption{}   
        \label{fig:esol_vis_1}
    \end{subfigure}
    \hfill
    \begin{subfigure}[b]{0.475\textwidth}  
        \centering 
        \includegraphics[width=\textwidth]{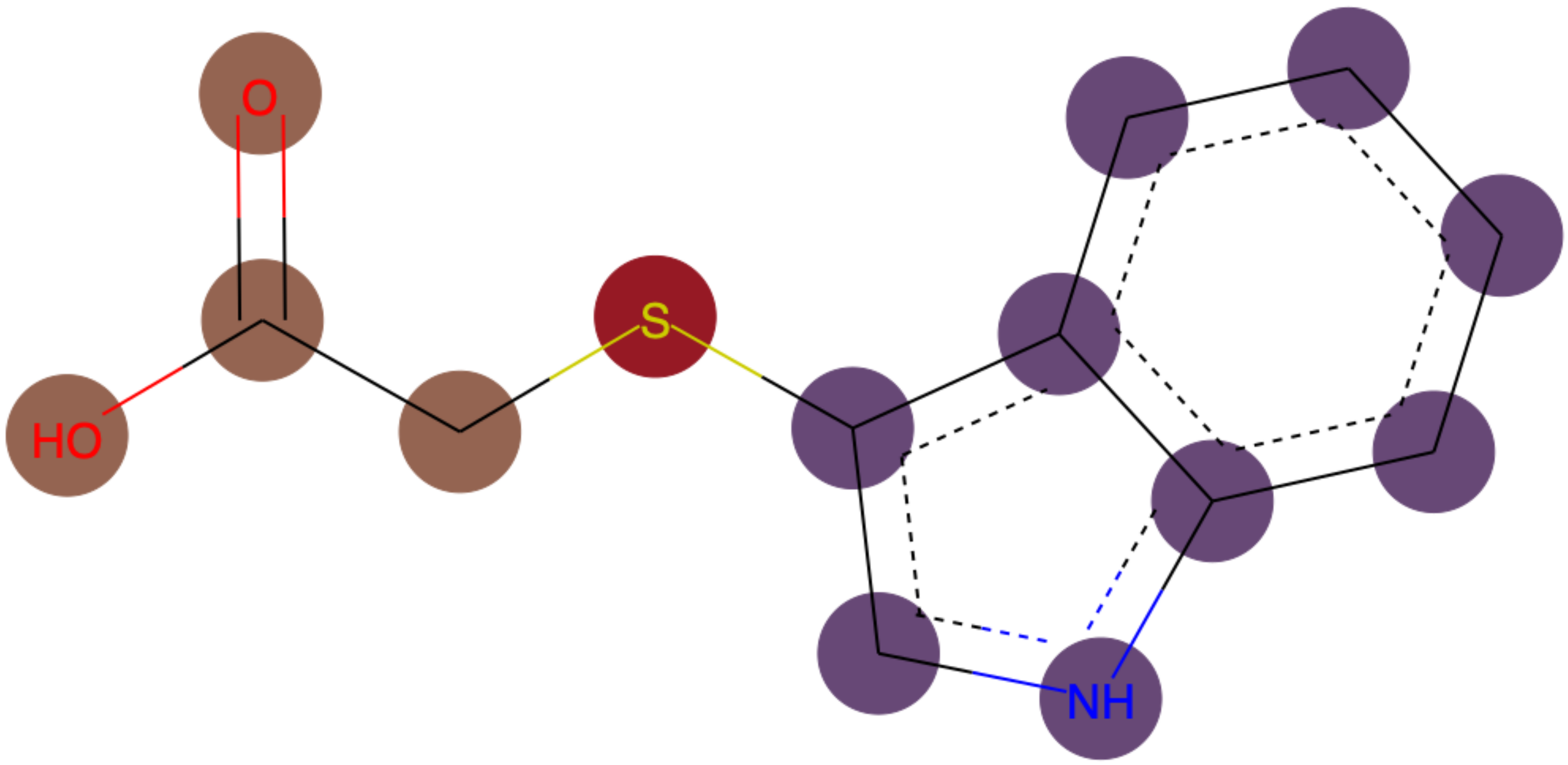}
        \caption{}    
        \label{fig:lipo_vis_1}
    \end{subfigure}
    \vskip\baselineskip
    \begin{subfigure}[b]{0.475\textwidth}   
        \centering 
        \includegraphics[width=\textwidth]{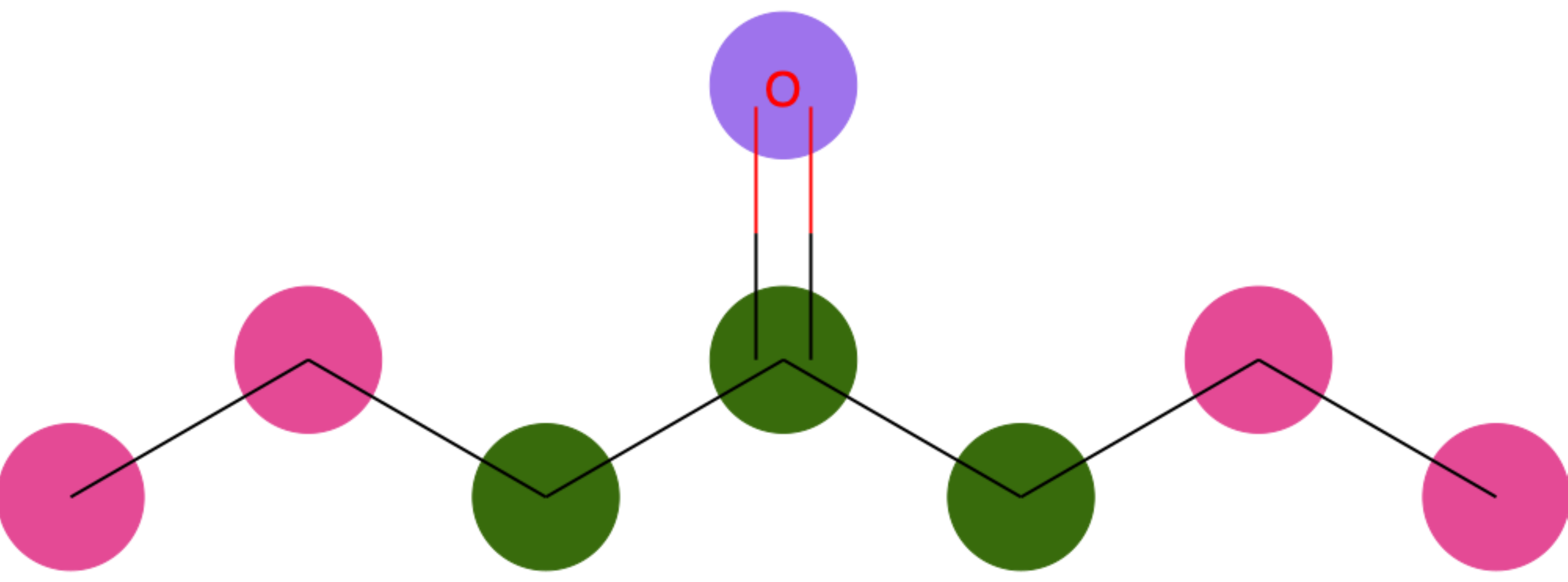}
        \caption{}    
        \label{fig:esol_vis_2}
    \end{subfigure}
    \quad
    \begin{subfigure}[b]{0.475\textwidth}   
        \centering 
        \includegraphics[width=\textwidth]{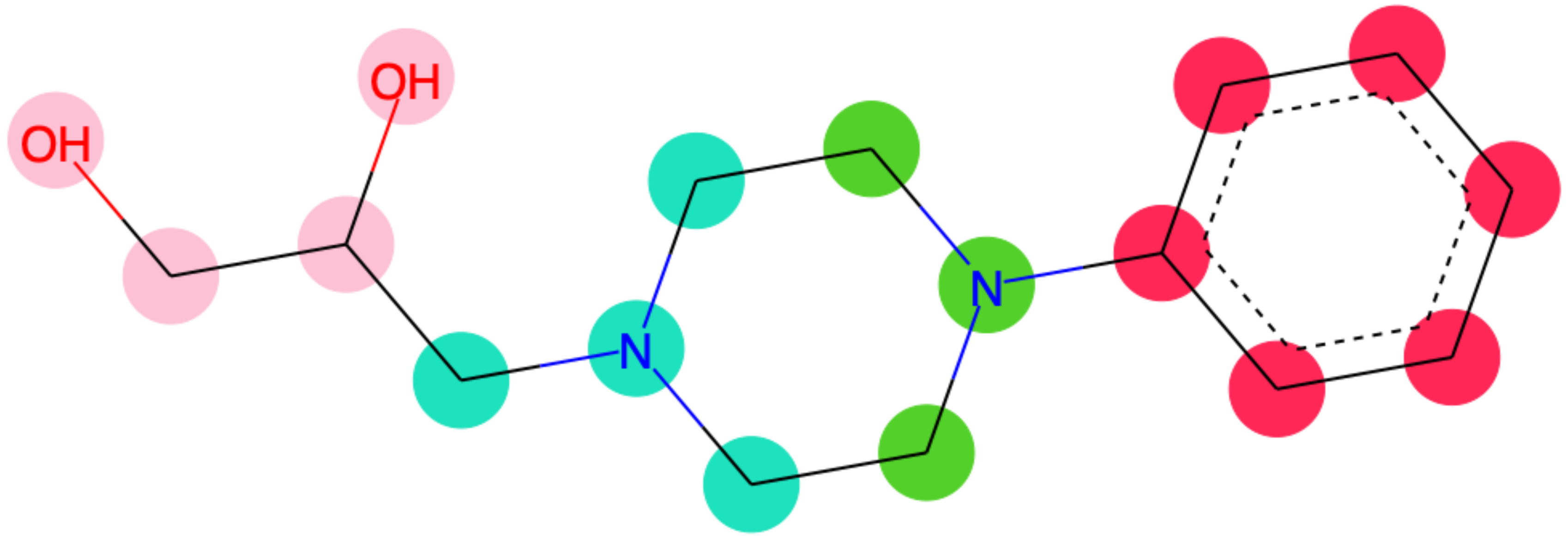}
        \caption{}    
        \label{fig:lipo_vis_2}
    \end{subfigure}
    \caption{Clusters learned by a MeMGNN for ESOL and LIPO dataset. Chemical groups like OH (hydroxyl group), CCl3, COOH (carboxyl group), CO (ketone group) as well as benzene rings have been recognized during the learning procedure. These chemical groups are highly active and have a great impact on the solubility of molecules.} 
    \label{fig:sample_vis}
\end{figure*}

Figure \ref{fig:C_F_comp} shows how unsupervised loss helps the model to push the nodes into distinct clusters. Figures \ref{fig:c_f_comp1} and\ref{fig:c_f_comp3} 
illustrates clusters with unsupervised loss and Figures \ref{fig:c_f_comp2} and \ref{fig:c_f_comp4} show computed clusters without unsupervised loss. The 
visualizations suggest that unsupervised loss helps the model to avoid trivial solutions by collapsing the latent node representations into meaningless clusters. Also, 
Figure \ref{fig:sample_vis} represents meaningful chemical groups extracted by MemGNN. Figures \ref{fig:lipo_vis_1} and \ref{fig:lipo_vis_2} are from LIPO and 
Figure \ref{fig:esol_vis_1} and \ref{fig:esol_vis_2} are from ESOL dataset respectively.
%------------------------------------------------------------------------------------------------------------------------------------------------------------------------------------------------------------------------
\begin{table}[t] \setlength{\tabcolsep}{5pt}
\begin{center}
\caption{Hyper-parameters selected for the models.}  \label{tab:hyper}
 \vspace{-0.2cm}
  \centering
  \begin{tabular}{lcccccc}
    \hline
    \textbf{Dataset} & \textbf{\#Keys} & \textbf{\#Heads} & \textbf{\#Layers} & $|$\textbf{Hidden Dimension}$|$ & $|$\textbf{Batch}$|$\\
     \hline
     Enzymes & [10,1]  & 5 & 2 & 100 & 20 \\
     Proteins & [10,1]  & 5 & 2 & 80 & 20 \\
     DD & [16, 8, 1] & 5 & 3 & 120 & 64\\
     Collab & [32, 8, 1] & 5 & 3 & 100 & 64\\
     Reddit-B & [32,1]  & 1 & 2  & 16 & 32 \\
     ESOL & [64,1]  & 5 & 2 & 16 & 32 \\
     Lipophilicity & [32,1]  & 5 & 2 & 16 & 32 \\
     BACE & [32,1]  & 5 & 2 & 8 & 32 \\
  \hline
  \end{tabular}
  \end{center}
\end{table}
%------------------------------------------------------------------------------------------------------------------------------------------------------------------------------------------------------------------------
\begin{table}
\addtolength{\tabcolsep}{-3pt}
\vspace{0.3cm}
  \caption{Summary of statistics of the benchmark dataset.}
  \vspace{-0.3cm}
  \label{tab:datastat}
  \centering
  \begin{tabular}{lccccccc}
    \hline
     \textbf{Name} & \textbf{Task} & \textbf{Graphs} & \textbf{Classes} & \textbf{Avg. Nodes} & \textbf{Avg. Edges}  & \textbf{Node Attr.} & \textbf{Edge Attr.}\\
    \hline
    Enzymes & classification & 600 & 6 & 32.63 & 62.14 & 18  & 0\\
    Proteins & classification & 1113 & 2 & 39.06 & 72.82 & 29 & 0\\
    DD & classification & 1178 & 2 & 284.32 & 715.66 & 0 & 0\\
    Collab & classification & 5000 & 3 & 74.49 & 2475.78 & 0  & 0\\
    Reddit-B & classification & 2000 & 2 & 429.63 & 497.75 & 0 & 0\\
    Bace & classification & 1513 & 2 & 34.09 & 36.86 & 32 & 7\\
    Tox21 & classification & 8014 & 2 & 17.87 & 18.50 & 32 & 7\\
    ESOL & regression & 1144 & - & 13.29 & 13.68 & 32  & 7\\
    Lipophilicity & regression & 4200 & - & 27.04 & 29.50 & 32  & 7\\
    \hline
  \end{tabular}
\end{table}
%------------------------------------------------------------------------------------------------------------------------------------------------------------------------------------------------------------------------

%------------------------------------------------------------------------------------------------------------------------------------------------------------------------------------------------------------------------
%------------------------------------------------------------------------------------------------------------------------------------------------------------------------------------------------------------------------
\end{document}